\documentclass[journal]{IEEEtran}
\usepackage{times}  %Required
\usepackage{helvet}  %Required
\usepackage{courier}  %Required
\usepackage{url}  %Required
\usepackage{graphicx}  %Required
\usepackage{amssymb}
\usepackage[numbers]{natbib}
\usepackage{amsmath}
\usepackage{caption}
\usepackage{subcaption}
\usepackage[ruled,vlined]{algorithm2e}
\usepackage{amsmath,amssymb,amsfonts}
\usepackage{algorithmic}

\ifCLASSINFOpdf
 
\else
 
\fi

\begin{document}
\title{Inverse Feature Learning: Feature learning based on Representation Learning of Error}


\author{Behzad Ghazanfari$^{\dagger}$, Fatemeh Afghah$^{\dagger}$, MohammadTaghi Hajiaghayi$^{\ddagger}$\\
%\{bg697,fatemeh.afghah\}@nau.edu\\
$^{\dagger}$ School of Informatics, Computing, and Cyber Security, Northern Arizona University, Flagstaff, AZ 86001, USA. \\
%taylorm@eecs.wsu.edu\\
$^{\ddagger}$ University of Maryland, College Park, MD 20742, USA.\\
}


\maketitle

\begin{abstract}
This paper proposes \textit{inverse feature learning} as a novel supervised feature learning technique that learns a set of high-level features for classification based on an \textit{error representation} approach. The key contribution of this method is to learn the representation of error as high-level features, while current representation learning methods interpret error by loss functions which are obtained as a function of differences between the true labels and the predicted ones. One advantage of such learning method is that the learned features for each class are independent of learned features for other classes; therefore, this method can learn simultaneously meaning that it can learn new classes without retraining. Error representation learning can also help with generalization and reduce the chance of over-fitting by adding a set of impactful features to the original data set which capture the relationships between each instance and different classes through an error generation and analysis process. This method can be particularly effective in data sets, where the instances of each class have diverse feature representations or the ones with imbalanced classes. The experimental results show that the proposed method results in significantly better performance compared to the state-of-the-art classification techniques for several popular data sets. We hope this paper can open a new path to utilize the proposed perspective of error representation learning in different feature learning domains.
\end{abstract}


\IEEEpeerreviewmaketitle

\section{Introduction}
\noindent Recent feature learning trend and its branches such as deep learning have offered remarkable performance in image, speech, and natural language processing.
Supervised or unsupervised representation learning are generally based on elements such as restricted Boltzmann machines (RBMs) \cite{smolensky1986info}, autoencoder \cite{bourlard1988auto,hinton1994autoencoders}, convolutional neural networks (ConvNets) \cite{lecun1998gradient}, sparse coding \cite{olshausen1997sparse,lee2007efficient}, and clustering methods \cite{lee2009convolutional,coates2012learning,xie2016unsupervised}. In the majority of existing supervised learning and representation learning methods, the term error refers to a function of the differences between the true and the predicted labels (i.e., loss functions). The error is utilized to optimize the training process or the learned features (e.g., optimizing the weights of neural nets). However, the notion of error can be considered in a more general term as a dynamic quantity that can capture the relationships between the instances and the predicted labels. Here, we develop a new framework for error generation and analysis called as \textit{error representation learning } to learn an additional set of impactful high-level features.

In this paper, we introduce the concept of error representation learning and propose inverse feature learning as a feature learning method based on error representation. This method is inspired by human's decision-making process that involves analysis and inference of results of its decisions. We believe proper error analysis to interpret error in the form of high-level knowledge is one of the missing puzzle pieces in the literature of feature learning. Such a perspective is somehow inspired by inverse reinforcement learning \cite{ng2000algorithms}, which attempts to learn the reward function instead of the optimal policies.
The idea behind the proposed inverse feature learning method as a supervised feature learning technique is to learn a new set of high-level features using a trial approach that investigates the impact of assigning the instances to different labels. This inverse feature learning method interprets error representation in the form of several variables  which depend on the predicted labels and the instances rather than the traditional notion of scalar values (e.g. loss functions). 

To the best of our knowledge, the proposed method is the first work that performs feature leaning based on the perception of error representation. This method proposes a new framework of learning by analyzing the interactions of instances and classes in a trail approach. During this trail phase, all possible labels are assigned to a test instance in order to generate the required perspective between the instances and each label in the form of new features. The key motivation of this paper is to introduce a new perspective for error representation in representation learning, and this basic proposed model can be improved in several levels and be applied in other domains. 

\section{Related Works} \label{Background}
The current methods in supervised or unsupervised representation learning are developed based on the representation of data. Some of the popular representation learning methods include ConvNets, RBMs, autoencoder, clustering methods, and sparse coding \cite{smolensky1986info,bourlard1988auto,hinton1994autoencoders,lecun1998gradient, olshausen1997sparse,lee2007efficient, lee2009convolutional,coates2012learning,xie2016unsupervised}. Supervised representation learning typically refers to a class of feature learning methods where the features are learned using the labeled data in a closed-loop manner. Deep learning techniques, as a subcategory of representation learning, learn a set of compact and high-level features of each class through multiple layers.

In recent representation learning methods, the unsupervised feature learning is usually used for pre-training nets, generally for deep networks, or for extracting high-level features, denoising, and dimensionality reductions. The authors in \cite{bengio2007greedy} used unsupervised learning for pre-training deep supervised networks, deep belief networks. Convolutional deep belief networks have been used for audio data in \cite{lee2009unsupervised} and image processing in \cite{lee2009convolutional}. The authors in \cite{coates2012learning} used K-means as clustering for each layer sequentially in a bottom-up approach and in an end-to-end way in \cite{caron2018deep}.

\begin{figure*}
\centering
\includegraphics[width=1\linewidth,height=6cm]{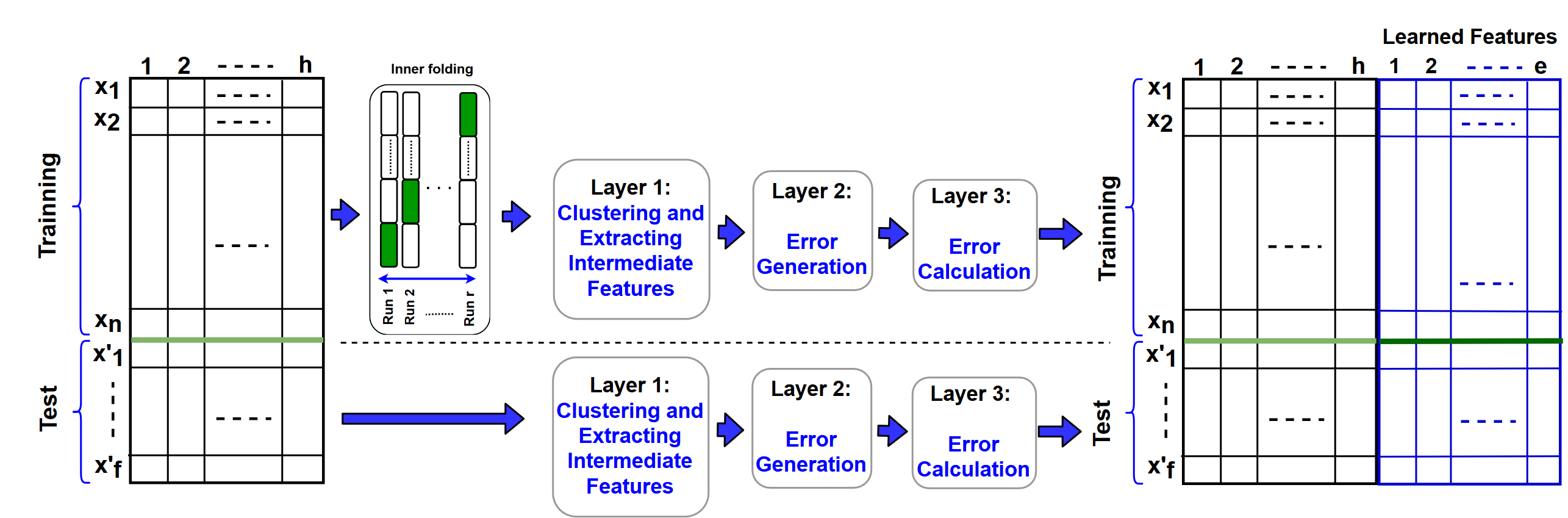} 
    \caption{Block diagram of the proposed inverse feature learning method. The figure demonstrates the feature learning process for both training and test sets. The upper part of the figure depicts the inner-folding for the training set during $r$ rounds. The one fold of the inner-test in each run is highlighted by green. The lower part of the figure demonstrates the feature learning process for the test data set. }
    \label{fig:diagram}
    %\vspace{-0.6 cm}
\end{figure*}

Here, we would like to point out the distinctions of the proposed inverse feature learning techniques related to some common trends in machine learning. The proposed error representation learning method learns high-level features depending on the instances and classes. This inverse feature learning method generates error by trial and calculates the resultant representation of the error. The proposed method is considerably different from existing techniques in the literature since they focus on data representation learning and calculate errors by the loss functions that take the difference between the predicted labels and true labels. 

This proposed method is different from generative adversarial nets (GANs) \cite{goodfellow2014generative}, which use two separate neural networks competing against each other with the ultimate goal of generating synthetic data similar to the original data inputs through the generator. This proposed method is also different from similarity learning \cite{koestinger2012large}, which learns a similarity function to measure how similar two objects are, in the sense that it extracts the relationships between the objects and the classes. Self-supervised learning methods are based on finding patterns inside of input instances \cite{de1994learning}.

The semi-supervised learning and active learning methods that utilize a combination of classification and clustering are based on the assumption that the instances which are in the same cluster have the same label and using this assumption toward predicting the labels for new instances  \cite{chapelle2009semi,benabdeslem2014efficient,seeger2000learning,chapelle2003cluster}. In these methods, the instances near the center of clusters are considered as the most representative objects to determine the labels. Other approaches including \cite{xu2003representative,nguyen2004active} utilized clustering for active learning in several different ways. %Some of the methods that have combined clustering with classification are in the form of ensemble learning \cite{evans2011clustering,ao2014combining}. EC3 \cite{chakraborty2017ec3} is an ensemble learning as a combination of several clustering and several classification methods by using an optimization function. 
The proposed inverse feature learning technique utilizes clustering for error representation learning in an innovative framework. Error representation learning is based on error generation and analysis scheme in which the instances are included in different classes and then clustered to different clusters in order to observe the relationships between the inserted instances and different members of each class and the caused effects of that insertion. 

Metric learning, similarity learning, techniques in semi-supervised learning, and adversarial autoencoders like GANs learning are based on data representation learning with the same traditional notion of error that refers to calculating the differences between the true labels and the predicted ones. However, our proposed method learns the representation of error that is intentionally generated in a trail process to generate an error for each class and to process this novel notion of class-dependent error in the form of features. The aforementioned methods do not generate and process representation of error for each and all classes simultaneously as a reference but rather use the error as a by-product of true and predicted labels. Therefore, our method develops a novel concept for error representation.

%More details about other related works such as generative adversarial nets \cite{goodfellow2014generative}, similarity learning \cite{koestinger2012large}, self-supervised learning \cite{de1994learning}, feature construction and extraction approaches \cite{guyon06,duda2012pattern, storcheus2015survey} and their usage in deep representation methods \cite{coates2012learning,kenyon2018clustering}, semi-supervised learning, active learning, as well as a review of the methods which used clustering in supervised learning are reported in ``supplementary material'' section along with a discussion to highlight the distinguished points and contribution of the proposed method compared to these aforementioned methods. 

%Unsupervised learning methods can be utilized for dimensionality reduction, pre-training, noise reduction in supervised learning methods. Clustering methods can enhance the performance of classification techniques (even for shallow networks) or other machine learning methods such as feature construction and extraction approaches \cite{guyon06,duda2012pattern}. Feature construction and extraction methods transform the input data instances to a number of features \cite{guyon06,storcheus2015survey}. For example, the centers of clusters can be used as meta-data points or representative of instances inside a cluster \cite{coates2012learning,kenyon2018clustering}.

\section{Inverse Feature Learning}\label{proposed_method}

In this section, we introduce the proposed inverse feature learning mechanism that learns the representation of error using a trial approach. The operation of this method for training and test instances is described in the following sub-sections \ref{subsec:train} and \ref{subsec:test}, respectively. The overall block-diagram of this method is shown in Figure \ref{fig:diagram}.

\subsection{Error Representation Learning for Training Instances} \label{subsec:train}
The objective of the inverse feature learning method is to learn a set of additional features per sample by trial to extract the relations between the sample and the classes. This process is performed during two phases for the training and test sets. In each phase, the samples are assigned to the \textit{set of samples of the available classes} one at a time and the changes in the characteristics of data before and after adding each sample are analyzed. Here, we provide an overview of the proposed method.

\begin{figure*}
\centering
\includegraphics[width=0.9\linewidth,height=8cm]{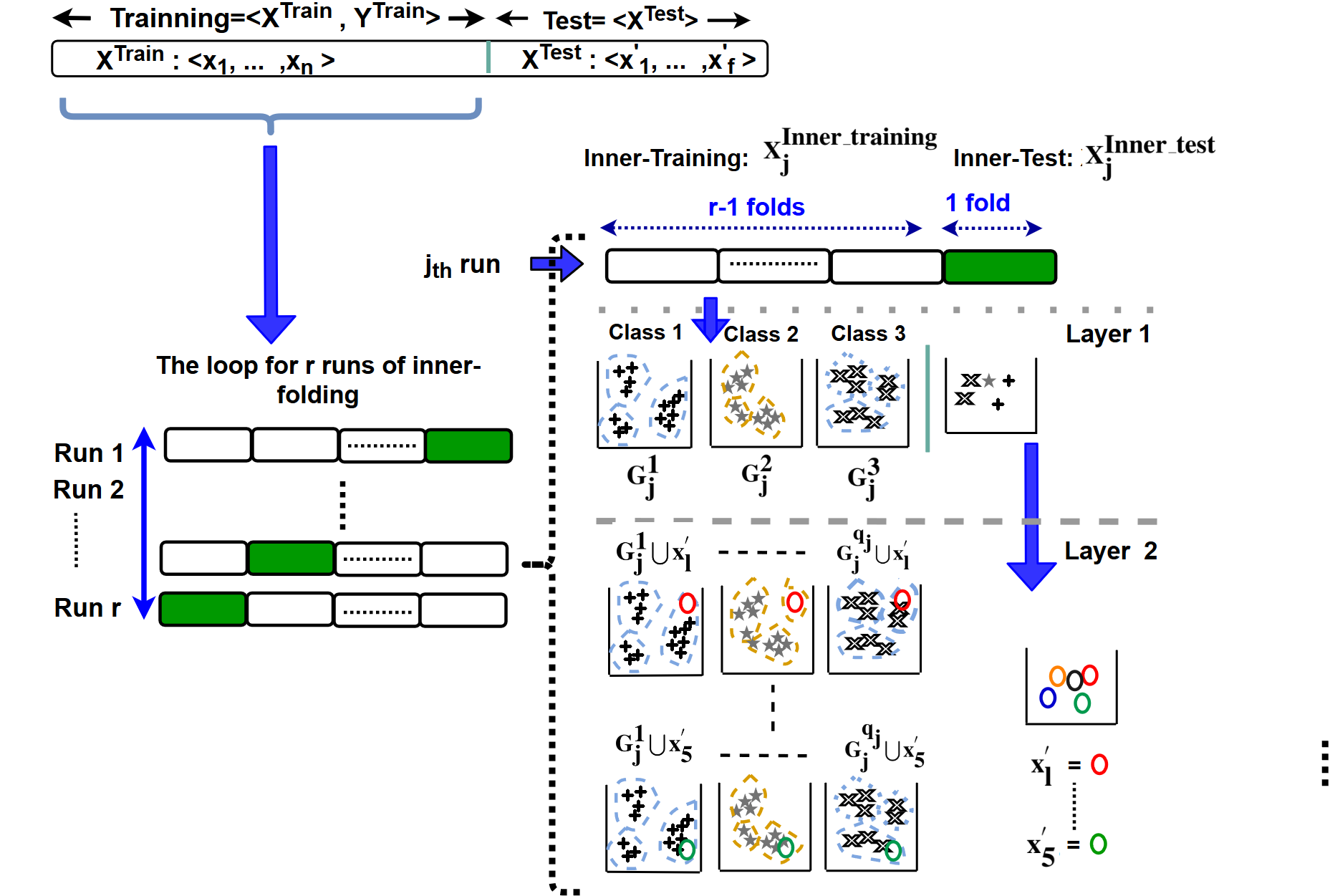}
    \caption{Diagram of the main blocks of the proposed feature learning process for the training data set including the inner-folding, layer 1 (clustering) and layer 2 (adding the inner-test samples). In this example, the number of classes in round $j$ is assumed as 3 (i.e., $q_j=3$). It is also assumed that the one fold inner-test includes five samples. These samples are denoted with circles in layer 2.} 
    \label{fig:fig1}
   % \vspace{-0.6 cm}
\end{figure*}

First, in inner folding phase, we partition the training instances to inner-training and inner-test sets during each fold in such a way that each training instance is considered as inner-test instance once. Then, in layer 1 of the proposed inverse feature learning, the inner-training samples with the same labels are grouped together. Next, the groups of samples (i.e., the samples with the same label) are clustered to a pre-determined number of clusters. The representation of these clusters for each label are calculated in the form of several intermediate features. 
In layer 2, each inner-test sample is intentionally assigned to all available classes, and then one of two described strategies are performed to extract and analyze a notation of error as a means to learn a new set of high-level features. In layer 3, regardless of the fact that the sample has been assigned to the right or wrong classes, we measure two sets of metrics per sample for each class. These two sets of learned metrics are then added to the original data as the learned features. Since these features are learned per class, the features belonging to different classes need to be separated from one another. Therefore, we embed the corresponding classes of these feature sets in them by multiplying the corresponding features of each class to a different large number. The model is trained and evaluated by adding the learned features per instance to the set of the primary features of the corresponding instance for both of training and test instances, respectively. In the feature learning for training instances, the features are only learned for inner-test samples, in which their labels are not considered in the process. The reason is because  we aim to develop a unified framework for feature learning during the training and test phases in classification, where the test instances do not have labels.

To formulate the problem, the input training data set is presented with $D=\langle X^{Train}, Y^{Train} \rangle$, in which  $X^{Train}=\{x_{1},\cdots, x_{n}\}$  indicates the set of input training instances and $n$ shows the number of input instances in the training partition.
Each instance $x_{i}=\langle x_{i,1}  , \cdots,x_{i,h} \rangle$ consists of $h$ features. 
The label set is denoted by $Y^{Train}$, where $Y^{Train}=\{y_{1},\cdots, y_{n}\}$ is a vector corresponded with data set $X^{Train}$. Thus, $y_{i}$ shows the corresponding label for $x_{i}$. Since we focus on classification, the labels are categorical. $\mathcal{Z}$= $\langle z_{1}, \cdots, z_{m} \rangle$ shows the set of classes, in which  $m$ indicates the number of classes. $X^{\text{Test}}=\langle x^{'}_{1},\cdots,x^{'}_{f} \rangle$ denotes the test set in which $f$ refers to the number of test instances. Notation $|b|$ indicates the number of instances in set $b$. %A summary of symbols and their corresponding definitions is presented in Table 1 %\ref{table_notations} of ``supplementary material''. 
In continue, we describe the building blocks of this method with more details, as depicted in Figure \ref{fig:fig1}.

\subsection*{Inner Folding} \label{subsec:Simulated_Folding} 
The first step is to find the representation of samples belonging to different classes for both training and test sets. It is simple to obtain the representation of each class for the test samples as we can simply partition the training samples to different classes in order to obtain the representation of each class.
However, the equivalent process of partitioning is more computationally complex for training samples, since each sample of the training set should be considered against all remaining samples with a strategy similar to leave-one-out cross-validation. In each case, the number of remaining samples to learn the class representation is $n-1$, in which $n$ denotes the number of training instances. Hence, this process involves a large number of repetitions of the feature learning process (i.e., $n$ times) that is not scalable to large data sets. Therefore, we instead apply a folding mechanism, here called as \textit{inner folding} to only perform the feature learning process for a limited number of runs (i.e., $r$-runs, where $r \ll n$).  During each round of inner folding, the training samples are divided into two partitions of inner-training and inner-test samples. 

We introduce inner folding as a partitioning mechanism that works similar to $k$-fold cross-validation in terms of partitioning data, but the objective of this inner folding is different from typical cross-validation folding methods. Inner folding is a framework that partitions $\langle X^{\text{train}}, Y^{\text{train}} \rangle$ to $r$ folds. It runs $r$ times, wherein  each run, $r-1$ folds are used for training and 1 fold is used for test. Here, each fold is considered as a test fold only one time. The training and test partitions in each run are called as \textit{inner-training} and \textit{inner-test}, respectively. Thus, inner folding is different from cross-validation since it is used as a mechanism to evaluate the characteristics of one test sample of inner-test against the inner-training samples in each run. %Also, there are the several further processes on the folds to learn features of error representation. 

The inner-training and inner-test sets of the $j^{\text{th}}$-run of the inner folding are shown with $\langle X^{\text{Inner\_train}}_{j}, Y^{\text{Inner\_train}}_{j} \rangle$, and $\langle X^{\text{Inner\_test}}_{j} \rangle$, respectively.  Clearly,  $X^{\text{Train}} = X^{\text{Inner\_test}}_{j} \bigcup X^{\text{Inner\_train}}_{j}$ for each $j$. The trail process involves  assigning each test instance, $\forall x_{f}^{'}  \in X^{\text{Inner\_test}}_{j}$, to each available label, $z_i$, $z_i \in \mathcal{Z}$ that exists in $Y^{\text{Inner\_train}}_{j}$.  In continue, we describe the three layers of error representation for each run of the inner-folding process.

% The proposed inner folding enables feature learning based on unsupervised learning. %in can be considered as a branch of nested cross-validation. Nested cross-validation is a two level procedure for optimizing parameters or selecting the model in supervised learning.% During $k$-fold cross validation, a loop with $k$ runs is defined on the data to split it to $(k-1)$ folds for training and one fold for test.
%The $(k-1)$-folds for training are considered as a new set that are folded to $r$ parts during the inner folding level. %In the second level, $r$ runs is doing in which in each run $r-1$ folds are selected for training and 1 fold is for test. 

%During the outer $k$-fold, let us consider the $t^\text{th}$ round of folding where $t\in \{1,...,k\}$. In this round   $k-1$ folds of $k$ folds for training and one fold for test in $t^{\text{th}}$ run in the first level of the nested cross-validation are indicated as $\langle X^{tr}_{1,t}, Y^{tr}_{1,t} \rangle$ and $\langle X^{\text{te}}_{1,t} , Y^{\text{te}}_{1,t} \rangle $ correspondingly. 

%We propose simulated folding as a framework to learn error by simulating error. In other words, we create a simulation environment during the inner layer of  nested folding to evaluate the \textcolor{magenta}{ behavior} of each inner-test samples when being assigned to different classes.   %In other words, in each run of the second level a trial of each possible label in the training of the existing labels in training of that run is doing for each test instance. 

\begin{algorithm}
   \caption{: Clustering and Extracting Intermediate Features for the $j^{th}$ run.}
   \label{alg:layer_0}
\begin{algorithmic}[1]
   \STATE {\bfseries Input:}  $G_{j} = \langle G^{1}_{j}, \cdots, G^{q}_{j} \rangle$;  the number of clusters ($k$).
   \STATE {\bfseries Output:} Calculating the mean of group $G_j$, clustering each member of $G_j$ (i.e., $G^{i}_{j}$) into $k$ clusters, $\langle C^{1}_{G^{i}_{j}},\cdots ,C^{k}_{G^{i}_{j}} \rangle$, and calculating mean, centroid, and confidence for each cluster.
    \FOR{i=1: $q_j$}
    \STATE Cluster $G^{i}_{j}$ into $k$ clusters.
    \STATE $\forall$ $x_{t}$ $\in$ $G^{i}_{j}$ \textbf{:}
  
    \FOR{l=1:h}
    \STATE $\mu^{i}_{j,l}$ $\leftarrow$ $\text{average}$ $(x_{t,l})$, where $x_{t,l}$ denotes feature $l$ of sample $x_t$.
    \ENDFOR
  
    \STATE Sort the clusters based on the number of instances in them :  $C_{G^{i}_{j}}= \langle C^{1}_{G^{i}_{j}},\cdots ,C^{k}_{G^{i}_{j}} \rangle$.\\
    \FOR{a=1:k}
    \STATE $ \text{Centroid}\_C^{a}_{G^{i}_{j}} \leftarrow$ The center of cluster $C^{a}_{G^{i}_{j}}$. 
    \STATE  $\text{Confidence}\_C^{a}_{G^{i}_{j}} \leftarrow \frac{|C^{a}_{G^{i}_{j}}|}{|C_{G^{i}_{j}}|}$. 
    \STATE $\text{Mean}\_C^{a}_{G^{i}_{j}} \leftarrow mean(\forall x_{t} \in C^{a}_{G^{i}_{j}})$.
    \ENDFOR
    \ENDFOR
\end{algorithmic}
\end{algorithm}

\subsection*{Layer 1: Clustering and Extracting Intermediate Features} \label{subsec:phase_2}
As mentioned before, the proposed inverse feature learning method is designed based on the analysis of error representation. The error is measured using different metrics after assignment of the inner-test samples to different classes in order to investigate the variations in the relative relations among the samples. To do that, first, the instances of each class are clustered to a pre-determined number of clusters using $K$-means algorithm as an unsupervised learning method. K-means algorithm is selected as the clustering technique since it is very fast and can be scaled to high-dimensional data sets \cite{coates2011analysis}. However, we understand that $K$-means algorithm does not perform well in cases, where the instances have different densities or the instances are distributed in non-spherical forms. We should note that the proposed inverse feature learning method is generic and can be implemented with other clustering techniques.

% $\langle X^{Inner\_train,i}_{j}, Y^{Inner\_train,i}_{j} \rangle$. %It means $Y^{Inner\_train,i}_{j}$ as labels of all the instances that belong to $G^{i}_{j}$ have the same label. %The set of groups is shown with  $G=G_{1},\cdots,G_{q}$, in which $q$ shows the number of different labels in $Y^{tr}_{2,t,j}$.% It is clear that $q \leq m$ and  $\bigcup\limits_{i=1}^{q} G_{i} = X^{tr}_{2,t,j}$.  and each $G^{tr,i}_{2,t,j}$ is shown as .
%A trial means the inner-test instance considered as an instance of a class. Thus, in each run of inner folding,the number of  trial equals with the number of inner-test instances multiplied to number of different available classes in inner-training of that run. 
%Also, it provides the confidence of possibility of belonging each instance to a label. The clustering assigns instances into the clusters based on a similarity measure. %The center of each cluster is presented by $Center_C$.The formula of mean, the formula of probability.
%In this paper, we learn error representation by simulation and trial each decision, label, before and after  separately.
%\cite{macqueen1967} 
%Also, there is not any theoretical proof to find the optimal number of clusters in advance.  %For example, the number of clusters for each label is considered as a constant value.

During each round $j$ of the inner-folding process, the instances of $X^{Inner\_train}_{j}$ are first categorized based on their labels, $Y^{Inner\_train}_{j}$, to $G_{j}= \langle G^{1}_{j}, \cdots, G^{q_j}_{j} \rangle$, in which $q_j$ shows the number of labels in $Y^{Inner\_train}_{j}$. The corresponding input instances and labels of  group $i$ during  round $j$, $G^{i}_{j}$, are shown with $X^{Inner\_train,i}_{j}$.
Then, the samples of each class $i$, $G^{i}_{j}$, are clustered to $k$ clusters. These clusters are ordered based on the number of their member instances and shown with $C_{G^{i}_{j}}= \langle C^{1}_{G^{i}_{j}},\cdots ,C^{k}_{G^{i}_{j}} \rangle$.  
%The clustering assigns instances into the clusters .
%\begin{align}
%\label{eq:kmeans}
%O= \sum_{t=1}^{k} \sum_{l=1}^{s} \Vert x_{l}^{t} - \text{Centroid}_{t}  \Vert^{2} 
%\end{align}

During the clustering, each object is assigned to the cluster that has the nearest centroid as a mean of its instances based on a distance metric. The objective function of K-means is considered to find the $\text{centroid}_{1},...,\text{centroid}_{k}$ in order to minimize the objective function, $O$, as described in (\ref{eq:kmeans}). %\vspace{-5pt}
\begin{equation}
O= \sum_{t=1}^{k} \sum_{l=1}^{s} \Vert x_{l}^{t} - \text{Centroid}_{t}  \Vert^{2}
\label{eq:kmeans}
\end{equation}
where $s$ denotes the number of instances in  group $G_j^i$ with label $i$, and $x_l^{t}$ denotes the instance $x_l$ that belongs to cluster $t$. 

\textit{Mean-group} is a metric defined as the mean of each group, $G^{i}_{j}$. This metric, as described in algorithm \ref{alg:layer_0}, is a vector denoted by $\mu^{i}_{j}$  that its $l^{\text{th}}$-element is calculated as the average of  $l^{\text{th}}$ features of all instances that belong to this group. 
%in the ``supplementary material'' section. %\textcolor{green}{It should be noted the ``mean-group'' is based on the typical summation and division and is different of the ``mean'' that are used in following that calculated depending on similarity metrics. }

In order to evaluate the characteristics of the clusters of each class, we define three other metrics of \textit{confidence}, \textit{mean}, and \textit{centroid}, as described in  algorithm \ref{alg:layer_0}.  These measures act similar to kernel functions of ConvNets to extract the representation of each cluster. The \textit{centroid} indicates the center of a cluster that is obtained by the clustering method. The \textit{Confidence} is a singular scalar value and defined as the ratio of the number of instances of each cluster to the number of all instances of that class. In other words, the confidence metric is a membership value that shows the probability that an instance belongs to a particular class. The \textit{mean} is a vector with length $h$ that calculates the average of $l^{\text{th}}$ features of all instances that belong to a cluster. The calculation of these measures in layer 1 captures the baselines to learn the error as defined in layer 2.
%\[ \textit{confidence} (c_{z_{i},j}) = \frac{|c_{z_{i},j}|}{\sum_{a=1}^{k}|c_{z_{a},j}|} \]

%\[
%\textit{mean} (c_{z_{i},j}) = mean(\forall x_{b} \in c_{z_{i},j}) \]
%\[\textit{mean} (c_{z_{i},j}) = \frac{ \sum_{a=1}^{h}\forall x_{b} \in c_{z_{i},j}  x_{b,a} }{|c_{z_{i},j}|} \]
%For example considering the set $A= \langle x_{1}, \cdots, x_{f}\rangle$. $A$ is clustered to $k$ clusters $C=\langle c_{1}, \cdots, c_{k} \rangle$. $|c_i|$ shows the number of instance of each cluster and center 

%In other words, the representation of error are learned while there is true label for each instance.

% \sum_{i=1}^{u} x_{i}x_{i} {x_{1},\cdots, x_{u}}
% Note use of \abovespace and \belowspace to get reasonable spacing
% above and below tabular lines.
%'''

%\subsection{Layer 0: A Robust Representation of Instances of Each Label} \label{subsec:phase_3}

\subsection*{Layer 2: Error Generation }
\label{subsec:phase_3}
%\textcolor{green}{ I need to set that .The pseudo codes of the layer 1 and both strategies of layer 2 are presented in Algorithm.1, Algorithm.2, and Algorithm.3 in the Algorithms section of ``supplementary material'', respectively.}

In machine learning methods, error is the result of differences between the predicted output and true output that is measured by loss functions and used to train the model. In a simple case, the error is ``one'' when a predicted label is incorrect, and the error is ``zero'' when the predicted label is correct. In the proposed inverse feature learning method, instead of using this traditional notion of error, the error is measured based on resultant representation of assigning the instances to different classes in a trail approach. 

As mentioned earlier, the instances that belong to each group  $G^{i}_{j}$ of $G^{j}$= $\langle G^{1}_{j}, \cdots, G^{ q_{j}}_{j} \rangle$ have the same label.  
In layer 1, the representation of each class and its clusters were measured using several intermediate features (e.g. the mean of a class, and the centroid, the mean, and  the confidence of the clusters). The goal of layer 2 is to evaluate the changes in these intermediate features, and in a more general sense, the representation of each class using the formed clusters by adding the test samples of inner-test. In other words, we assign the samples of the inner-test set to the existing labels one at a time. Therefore, the term trail refers to the process of inserting a new inner-test sample, $x^{'}_{l} \in X^{Inner\_test}_{j}$, to a group of samples with the same label, $G_j^i$ and generating a new set of ($X^{Inner\_training,i}_{j} \bigcup x'_l $). 

We have considered two strategies to evaluate the characteristics of data after addition of new inner-test samples. In the first strategy, the similarity between the instance and the centers of clusters of each class is measured to find the closest one in order to assign that instance to this closest cluster. In other words, this sample is added to the closest cluster. Thus, in the first strategy, which is the simpler option, is to calculate the distances between the sample of inner-test, $x_{p}^{'}$, ($x_{p}^{'} \in X^{Inner_{{test}_j}}$) and the centers of all clusters in $C_{G_{j}}= \langle C^{1}_{G^{i}_{j}},\cdots ,C^{k}_{G^{i}_{j}} \rangle$ (obtained in algorithm  \ref{alg:layer_0}), and then assign $x_{p}^{'}$ to the closest cluster of the class $i$. This cluster is denoted by $C^{*}_{G^{i}_{j}}$. After that the confidence, centroid, and mean are only calculated for the nearest cluster. This strategy is described in Algorithm \ref{alg:layer_21}. 

In the second strategy, the formed clusters of layer 1, $C_{G^{i}_{j}}$, are no longer used. Instead, in the this strategy, the set of instances in each class including the primary and new instances are clustered again. 
The first strategy involves less computations, however, it does not necessarily perform as well as the the second strategy of re-clustering when the density of clusters are considerably different. For instance, when there exists several outliers in the original groups.

\begin{algorithm}
   \caption{: Error Generation using the first strategy.  }
   \label{alg:layer_21}
\begin{algorithmic}[1]
   \STATE {\bfseries Input:} $C_{G^{i}_{j}}=\langle C^{1}_{G^{i}_{j}}, \cdots, C^{k}_{G^{i}_{j}} \rangle$; $X^{Inner\_test}_{j}$; the number of clusters ($k$).
   \STATE {\bfseries Output:}  
   $C^{'*}_{G^{i}_{j}}$; $\mu^{'i}_{j}$; $\textit{Centroid}\_C^{'*}_{G^{i}_{j}}$; $\textit{Confidence}\_C^{'*}_{G^{i}_{j}}$; and $\textit{Mean}\_C^{'*}_{G^{i}_{j}}$. 
    \FOR{each $x^{'}_p\in  X^{Inner\_test}_{j}$}
    \STATE Find the closest cluster by measuring  the distance between $x^{'}_p $  and the center of each cluster, $\langle C^{1}_{G^{i}_{j}}, \cdots, C^{k}_{G^{i}_{j}} \rangle$. The closest cluster is denoted by $C^{'*}_{G^{i}_{j}}$.\\ %represent the resultant cluster of the set $C_{G^{i}_{j}} \bigcup x^{'}_p $in which index of the closest center by $*$.
    Sort the clusters based on their number of instances. 
    
    \STATE $\forall$ $x_{t}$ $\in$ $C^{'*}_{G^{i}_{j}} $ \textbf{:}
    
    \FOR{l=1:h}
    \STATE $\mu^{'i}_{j,l}$ $\leftarrow$ $\text{average}$ $(x_{t,l})$, where $x_{t,l}$ denotes feature $l$ of sample $x_t$.
    \ENDFOR
   % \STATE find the nearest centroid among $k$ clusters of $\langle C^{1}_{G^{tr,i}_{2,t,j}},\cdots ,C^{k}_{G^{tr,i}_{2,t,j}} \rangle$ and calculate 
    \STATE $ Centroid\_C^{'*}_{G^{i}_{j}} \leftarrow$ The center of cluster $C^{'*}_{G^{i}_{j}}$. 
    \STATE  $\textit{Confidence}\_C^{'*}_{G^{i}_{j}} \leftarrow \frac{|C^{'*}_{G^{i}_{j}}|}{|C^{'}_{G^{i}_{j}}|}$. 
    \STATE $\textit{Mean}\_C^{'*}_{G^{i}_{j}} \leftarrow mean(\forall x_{t} \in C^{'*}_{G^{i}_{j}})$.
    
    \ENDFOR
\end{algorithmic}
\end{algorithm}

\subsection*{Layer 3: Error Calculation }

The purpose of layer 1 was to capture the representation of classes in the form of clusters before the trial process. In layer 2, the representation of each inner-test instance and classes after the trial was captured when each inner-test instance is added  to a cluster of each class. Now in layer 3, the representation of error is learned through several high-level features by non-linear modules as described in following. These features capture the relationships between the inner-test samples and the clusters through two feature sets. The definition of these features are summarized in Algorithm \ref{alg:layer_3}.

\textbf{Feature Set 1: The distances between the new instance and the clusters of each label}\\
This feature set captures the representation of error inside of each class including the distance of this instance to its closest adjacent instance in the class ($feature_{1.0}$), the distance of the instance with the mean-group ($feature_{1.1}$) as well as the distance of the inner-test sample with the centers of all clusters for each class (feature set: $feature_{1.2}$), the distance of the inner-test sample with the means of all clusters of each class (feature set: $feature_{1.3}$). 

 %\textcolor{blue}{If $x{'}_{l} \in X^{Inner\_test}_{j}$, then the corresponding features are calculated of $x{'}_{l}$  for each cluster. In other words, a feature vector in the length of $k$ is formed  in which each element of the vector correspond to a cluster.}% Thus, for $a=1:k$ \(\) 

\textbf{Feature Set 2: The changes in class representations before and after inserting the new inner-test sample} 

To capture the effects of error in the level of classes, the distances between the means and confidences of the clusters of each class, formed in layer 1 and layer 2,  are calculated. Also, the distances between the centers and means of clusters that were formed in layer 1 and layer 2 within each class are measured. We also measure the difference between the confidence metric of original cluster formed in layer 1 and the cluster including the sample in layer 2 that represents the membership value denoted by $feature_{2.3}$.  

Each instance in the training is considered as the inner-test instance for one time and then these features are calculated for this sample. The learned features in layer 3 are added to the set of the primary features of that inner-test instance. Thus, the output of the inner folding is a set of features per instance of the training. Since these features are calculated per class, they need to be differentiated from each other. Therefore, the features of each class are multiplied by the class ID to be separated from each other. For example, if we have three classes, a, b, and c, the learned features of these classes are multiplied to 1, 10, and 20, respectively to be distinguished from each other. Such separation of learned features of different classes cannot be easily handled by some classification methods, therefore, we utilized ensemble decision tree as the classifier for our approach to best treat this proposed embedded distance mechanism. In the following section, the process of feature learning for the test instances $X^{Test}$ has been described.

\begin{algorithm}%[H]
   \caption{: Error Calculation.}
   \label{alg:layer_3}
\begin{algorithmic}[1]
   \STATE {\bfseries Input:}  $C_{G^{i}_{j}}$; $C^{'}_{G^{i}_{j}}$; $\mu^{i}_{j}$; $\mu^{'i}_{j}$; $\textit{Centroid}\_C_{G^{i}_{j}}$; $\textit{Centroid}\_C^{'}_{G^{i}_{j}}$; $\textit{Confidence}\_C_{G^{i}_{j}}$; $\textit{Confidence}\_C^{'}_{G^{i}_{j}}$; $\textit{Mean}\_C_{G^{i}_{j}}$; $\textit{Mean}\_C^{'}_{G^{i}_{j}}$; the number of clusters ($k$).
   \STATE {\bfseries Output:} $feature_1, \cdots, feature_e$ as the vector of learned features for each instance in $X^{Inner\_test}_{j}$.
    \FOR{each $x^{'}_{l} \in X^{Inner\_test}_{j}$}
    \STATE $========$\textbf{Feature set 1}$=========$
    \STATE $------------1.0$
    \STATE $ feature_{1.0} \leftarrow distance(x^{'}_{l},\ \text{the closest instance}\ in\ C^{'*}_{G^{i}_{j}}))$
    \STATE $------------1.1$
    \STATE $ feature_{1.1} \leftarrow distance(x^{'}_{l}, \mu^{'i}_{j})$
    
    \STATE $------------1.2$
    \FOR{a=1:k}
    \STATE $ feature^{a}_{1.2} \leftarrow distance(x^{'}_{l}, \text{Centroid}\_{C^{'a}_{{G^{i}_{j}}}})$
    \ENDFOR
    %\STATE $------------1.3$
    % \FOR{a=1:k}
    %     \STATE $ feature^{a}_{1.3} \leftarrow distance(x^{'}_{l}, \text{Confidence}\_C^{'a}_{G^{i}_{j}})$
    % \ENDFOR
     \STATE $------------1.3$
     \FOR{a=1:k}
         \STATE $ feature^{a}_{1.3} \leftarrow distance(x^{'}_{l}, \text{Mean}\_{C^{'a}_{G^{i}_{j}}})$
     \ENDFOR
     \STATE $========$\textbf{Feature set 2} $=========$
     \STATE $------------2.1$
     \STATE $ feature_{2.1} \leftarrow distance (\mu^{i}_{j}, \mu^{'i}_{j})$
    
    \STATE $------------2.2$
    \FOR{a=1:k}
    \STATE $ feature^{a}_{2.2} \leftarrow distance (\text{Centroid}\_{C^{a}_{{G^{i}_{j}}}}, \text{Centroid}\_{C^{'a}_{{G^{i}_{j}}}})$
    \ENDFOR
    \STATE $------------2.3$
     \FOR{a=1:k}
         \STATE $ feature^{a}_{2.3} \leftarrow distance(\text{Confidence}\_C^{a}_{G^{i}_{j}}, \text{Confidence}\_C^{'a}_{G^{i}_{j}})$
     \ENDFOR
     \STATE $------------2.4$
     \FOR{a=1:k}
         \STATE $ feature^{a}_{2.4} \leftarrow distance(\text{Mean}\_{C^{a}_{G^{i}_{j}}}, \text{Mean}\_{C^{'a}_{G^{i}_{j}}})$
     \ENDFOR

     \ENDFOR

\end{algorithmic}
\end{algorithm}

%\textbf{Feature Set 3: The distances between the clusters of each class that encounter the new inner-test sample. }
%The confidence of chosen cluster is considered as a feature.

%The distance between the instance and the center of the selected cluster of each label is calculated.

%The distance between the instance and the mean of the selected cluster of each label is calculated.
%\\-------------------------\\
%The distance between mean, center, and confidence of  the selected cluster and corresponding features of all other clusters in that class for each class.
%\\-------------------------\\
%The distance between mean, center, and confidence of  the selected cluster and corresponding features of all the selected clusters of other labels.
% \\-------------------------\\
%We trial by assigning each possible label to a instance to capture the behaviour of each assigning for that instance. 

\subsection{Error Representation Learning for Test Instances} \label{subsec:test}
In this phase, as is shown in Figure \ref{fig:diagram}, the entire training set along with its additional learned features can be used as the training instances in the feature learning method (i.e., the inner folding process is no longer required). Therefore, the training and test instances are fed to the three-layer feature learning method to learn corresponding features for the test instances. We would like to note that the training instances are grouped based on their labels and then clustered. The test instances, which their labels are not available, are assigned to the proper cluster of each label and then the features are learned for each test instance. Hence, this step does not require the labels of test instances. Finally, the extended training and test data sets with the learned features are fed to the classifier.

\section{Time Complexity}\label{sec:Time_Complexity}
The proposed inverse feature learning method is developed based on clustering the data set using $K$-means algorithm through a trial process to learn several high-level features based on representation of error. For the sake of simplicity, this process is explained in a sequential way (algorithm \ref{alg:layer_0}), however, the process of clustering for different instances can be performed simultaneously. Therefore, the time complexity of this method would be a function of time complexity of the underlying $K$-means clustering algorithm.  

The time complexity of Lloyd's algorithm of $K$-means is ${O}(eknh)$, in which $e$ is the number of iteration, $k$ is the number of clusters, $n$ is the number of instances, and $h$ is the number of features \cite{lloyd1982least}. Since the number of iterations, $e$, the number of features, $h$, and the number of clusters, $k$, are constant, the time complexity of this algorithm is linear. In the second strategy of error generation, the samples in each class are clustered again after adding a new test sample as described in Section \ref{subsec:phase_3}. Hence, the $K$-means clustering method is performed $2\times{r}$ times during the inner folding process, in which the number of runs, $r \ll n$, is a constant number. Thus, the time complexity of the feature learning process is linear. Obviously, the time complexity of the first strategy in error generation, where the new instance is added to the cluster with the closest center is also linear.

\section{Experimental Results} \label{sec:results}

In this section, we evaluate the performance of the proposed inverse feature learning method using several popular binary and multi-class data sets. To do so, we compared its performance versus several classification methods that only use the original features. We embed the corresponding classes of learned feature sets of instances by multiplying the corresponding features of each class to a different number. All of those features of different classes are added to the corresponding original set of features of instances. Although, this multiplication and adding the learned features per class to the original ones may worsen the performance of some classifiers as they are required to deal with a bigger set of features including the original and learned features. Thus, the classifiers that consider the whole features as a unified set are subject to a weak performance here. Therefore, we used boosting decision trees as the main classifier for the proposed method because it works based on several decision trees that can better handle a bigger set of features of learned and original features corresponding to different classes compared to other classifiers. Different metrics including accuracy and F1 scores are used to evaluate the performance of our method. Also, we compared the performance of this inverse feature learning versus several deep representation learning approaches, as described in \cite{yu2015learning}, such as Linear ELM \cite{huang2004extreme}, Deep Belief Networks \cite{hinton2006reducing}, Stacked Auto-Encoder \cite{bengio2007greedy} for pre-training weights of the deep network alongside a softmax classifier \cite{greene2003econometric}, DrELM \cite{yu2015learning}, and $DrELM^{r}$ \cite{yu2015learning}. The results are reported using ``Statistics and Machine Learning Toolbox" of MATLAB {\textregistered}.

%\section{Theoretical Proof}
%The representation of error is the  minimization of error in assigning inner-test instances to each class satisfy by the cluster description of that class and adding each instance to the the nearest cluster. In fact, the clustering representation make to have much more accurate assignment, and capturing possibilities. It guarantee accurate analyze about corresponding error and the confidence of each assumption.

\begin{table}
\vspace{-10pt}
\caption{The specification of data sets.}
\label{datasetsdesriptoin}
\vspace{-10pt}
\begin{center}
\begin{small}
\begin{sc}
\scalebox{0.9}{
\begin{tabular}{lccc}%|cccr
%\toprule
\textbf{Data sets} & \textbf{\#instances} & \textbf{\#features} & \textbf{\#class}\\ %& Data sets & \#inst. & \#feat. & \#class \\
%\midrule
\hline
Cryotherapy    & 90 &  7& 2 \\
\hline
Diabetes    & 768 &  8& 2 \\
\hline
Heart    & 270 &  13& 2 \\
\hline 
credit   & 30000 &  23& 2 \\
\hline
Segment    & 1500 &  19& 7 \\
\hline
Ionosphere    & 351 &  34& 2 \\
\hline
Glass    & 214 &  9& 6 \\
\hline 
Spam & 4601 &57& 2\\
\hline
magic    & 19020 &  10& 2  \\
%\bottomrule
\end{tabular}}
\end{sc}
\end{small}
\end{center}
%\vskip -0.1in
\vspace{-12pt}
\end{table}

\begin{table*}[t]
\caption{The comparison of accuracy and F1 score of the baseline classifiers and the inverse feature learning. We have evaluated the performance of data sets with several popular classifiers libraries in MATLAB and sklearn including SVMs and logistic regressing but only the best results among different classifiers are reported. In the proposed method, the learned features are independent of the classifier, hence using a better classifier can achieve higher accuracy over our learned features as it does over other sets of features.}
\label{table:datasets_result}
\vskip 0.15in
\begin{center}
\begin{small}
\begin{sc}
\resizebox{2\columnwidth}{!}{
\begin{tabular}{ll|cc|cc|cc|cc|cc}%{lcccccccccc}%{l|cc|cc|cc|cc|cc}
%\firsthline
\hline
\# &Data sets & \multicolumn{2}{c}{Naive Bayes}&  \multicolumn{2}{c}{KNN}& \multicolumn{2}{c}{Decision Tree}& \multicolumn{2}{c}{Ensemble}& \multicolumn{2}{c}{Inverse Feature learning}\\
\hline
& & F1 & Accuracy & F1 & Accuracy & F1 & Accuracy & F1 & Accuracy & F1 & Accuracy \\
\hline
1& Cryotherapy  & 0.9109 & 0.9098  & 0.8994 & 0.9054 & 0.9221& 0.9321 & 0.8656 & 0.8791 & \textbf{0.9666}& \textbf{0.9667}\\
2& Heart    & 0.7910 & 0.7864  & 0.7975 & 0.8115 & 0.7627& 0.7827  & 0.7773 & 0.7890 & \textbf{0.8567}& \textbf{0.8593}  \\
3 & Segment    & 0.8898 & 0.8844  & 0.9607 & 0.9600 & 0.9393& 0.9439 & 0.9762 & 0.9761 &\textbf{0.9804}& \textbf{0.9807}  \\
4& Glass   & 0.5396 & 0.6155  & 0.7477 & 0.7512 & 0.6571& 0.6925 & 0.7085 & 0.7470 & \textbf{0.7809}& \textbf{0.7850} \\
5& magic     & 0.6907 & 0.7603  & 0.8013 & 0.8244 & 0.8084& 0.8214 & 0.8549 & 0.8699 &\textbf{0.8597} & \textbf{0.8753} \\
6&Diabetes   & 0.6797 & 0.7193  & 0.6569 & 0.6950 & 0.6825& 0.7118 & 0.6883 & 0.7181 & \textbf{0.7456}& \textbf{0.7721}  \\
7& Ionosphere     & 0.9019 & 0.8961  & 0.8956 & 0.9011 & 0.8691& 0.8835 & 0.9243 & 0.9257 & \textbf{0.9564} & \textbf{0.9601}   \\
8& Spam     & 0.5147 & 0.5241  & 0.9129 & 0.9184 & 0.9127& 0.9186 & 0.9464 & 0.9552 & \textbf{0.9562} & \textbf{0.9583}    \\
9& credit   & 0.4979 & 0.7193  & 0.6110 & 0.7276 & 0.6238& 0.7394 & 0.6714 & 0.8072 & \textbf{0.6783}& \textbf{0.8109} \\
\hline
\end{tabular}
}
\end{sc}
\end{small}
\end{center}
\vskip -0.1in
\end{table*} 

\begin{table}
\caption{ Comparison of the performance of the proposed inverse feature learning method with the reported results in \cite{yu2015learning} in terms of accuracy for two data sets of Segment and Diabetes. The reported results of other methods were tuned to provide their best performance.}
\label{DrELM}
%\vskip 0.15in
\begin{center}
\begin{small}
\begin{sc}
%\resizebox{1\columnwidth}{!}{
\begin{tabular}{lcr}
\hline
Data sets & Segment & Diabetes  \\
\hline
Linear ELM \cite{huang2004extreme} & 0.8661 & 0.7459\\
Deep Belief Networks \cite{hinton2006reducing} & 0.9551 & 0.7805\\
Stacked Auto-Encoder \cite{bengio2007greedy} & 0.9568 & 0.7783\\
DrELM  & 0.9530 &  0.7763  \\
$DrELM^{r}$    & 0.9579 & \textbf{0.7822} \\
Inverse Feature Learning &  \textbf{0.9807}&  0.7721\\
\hline
\end{tabular}%}
\end{sc}
\end{small}
\end{center}
\vskip -0.1in
\end{table}

We evaluated the results of our method over several runs for different data sets and the results showed very small variances. Theoretically speaking, since we just added a number of features and used the common classifiers that do not depend on stochastic behaviors; therefore, the results only have very small variances over different runs. The only stochastic part of the method is the clustering, where the clustering methods such as k-means can be easily stabilized   \cite{kuncheva2006evaluation, rakhlin2007stability} and are used in feature learning methods \cite{coates2012learning, xie2016unsupervised}.

A number of common data sets with a different number of instances, features, and classes are used in this study including  Cryotherapy, Heart, Segment, Magic, Letter, Credit, Spam, and Ionosphere \cite{Dua:2017}. The characteristics of these data sets are summarized in Table \ref{datasetsdesriptoin}. The reported results are obtained using $k$-fold cross-validation. Let us first describe the parameters involved in the proposed method before presenting the results.  %\colorbox{yellow}{lower k has been used for clustering!} The number of folds for cross-validation is selected as 5 or 10. 

% \begin{table}[t]
% \vspace{-10pt}
% \caption{The specification of data sets. \textcolor{magenta}{break this to a two-column one to save space}}
% \label{datasetsdesriptoin}
% %\vskip 0.15in
% \begin{center}
% \begin{small}
% \begin{sc}
% \begin{tabular}{lcccr}
% \toprule
% Data sets & \#instances & \#features & \#class \\
% \midrule
% Cryotherapy    & 90 &  7& 2  \\
% Heart    & 270 &  13& 2  \\
% Segment    & 1500 &  19& 7  \\
% Glass    & 214 &  9& 6  \\
% magic    & 19020 &  10& 2  \\
% Diabetes    & 768 &  8& 2  \\
% credit   & 30000 &  23& 2  \\
% Ionosphere    & 351 &  34& 2  \\
% Spam & 4601 &57& 2\\
% \bottomrule
% \end{tabular}
% \end{sc}
% \end{small}
% \end{center}
% %\vskip -0.1in
% \vspace{-12pt}
% \end{table}

As described in Section \ref{proposed_method}, the feature learning process for training data is based on a $r$-fold inner folding process, where $r$ shows the number of folds. Each round of inner folding involves clustering the training data set using $K$-means, in which $k$ is the number of clusters. The clustering can be performed using various distance metrics such as Minkowski, Euclidean, Cosine, Jaccard, City Block among other distance metrics. 
% \ref{table:parameters}
The corresponding parameters used in the proposed inverse feature learning method for the results in Table \ref{table:parameters} are described. The number of folds for cross-validation of the baseline classifier in Table \ref{table:datasets_result} is selected as 5 or 10 as used for the proposed method in the second column of Table \ref{table:parameters}.

The reported results are based on using the first strategy of error generation (i.e., assigning the test instances to the closest cluster). The results are reported for the case that the proposed inverse feature learning method is performed over boosting decision trees classifier. The reason behind selecting the boosting decision trees is its robust performance versus different feature sets per class since we embedded the label of each class in the learned features of that class.  Different baseline classifiers such as Naive Bayes (NB), Decision Tree, $K$-nearest neighbors (KNNs), and an Ensemble Classifier in form of boosting decision trees, as the same classifier that we used for inverse feature learning, are used for the sake of comparison. 
%and the inverse feature learning can be used alongside them.

As it can be seen in Table \ref{table:datasets_result}, the proposed feature learning method  provides considerably better results in different data sets compared to the known classifiers that only work with original features. It means that the learned features using our method can significantly improve the results of popular classifiers. The parameters of the proposed feature learning method are fine-tuned based on the data sets as described in Table \ref{table:parameters}. In Table \ref{table:parameters}, the third column describes the set of learned features (as described in Algorithm \ref{alg:layer_3}) that are used for each dataset. The fourth column of this table describes the underlying distance metric used for clustering. In this table, $L$ refers to the layer, and its following number refers to the levels of layers. For example, $L1$ refers to layer 1. The distance metrics are abbreviated. For example, `CB' refers to City Block, `JA' stands for Jaccard and `EU' represents Euclidean distance.

\begin{table}
\caption{The description of parameters used in the proposed inverse feature learning method. }
\label{table:parameters}
%\vskip 0.15in
\begin{center}
\begin{small}
\begin{sc}
\resizebox{1\columnwidth}{!}{
\begin{tabular}{lccccc}
\hline
Data sets & k-fold & $r$ & $k$ & feature set & distance metrics\\
\hline
Cryotherapy    & 10 & 5& 5 & 2.2 & L1,3:CB, L2:Ja  \\
Heart    & 10 &  2& 3 & 1.0, 1.2, 1.3& L1:Eu, L2: CB \\
Segment  & 10 &  5& 5 & 1.0, 1.1, 2.1, 2.4 & L1,3:CB, L2:Ja \\
Glass    & 5 & 5& 3 & 1.1& L1:EU, L2,3:Ja \\
magic    & 10 &  5& 10 & 2.1, 2.4 & L1,3:Eu, L2:Ja\\
Diabetes & 5 & 2& 10 & 2.1, 2.4 & L1,3:Eu, L2:Ja\\
credit   & 10 &  5& 10 & 1.0, 1.1 & L1,3:CB, L2:Ja \\
Ionosphere    & 5 &  5&5 & 1.0 & L1:CB, L2:Ja, L3:EU \\
Spam & 10 &3& 15&  2.3  & L1:EU, L2,3: Ja\\
\hline
\end{tabular}}
\end{sc}
\end{small}
\end{center}
\vskip -0.1in
\end{table}

We also compared the result of the proposed method in terms of accuracy with several most known approaches that learn deep representation in Table \ref{DrELM}. As it can been seen in the table, the proposed method can provide comparable results with these methods or outperform for some data sets, while it is worth noting that our method can learn new classes in an incremental manner without modifying the learned features of previous classes. However, the deep representation learning generally cannot be easily scaled-up to learn new classes without re-training. Based on the well-known ``no free lunch" theorem, there is no method that is superior to other methods over all data sets. As it can be seen in Table \ref{DrELM}, for Diabetes data set, the accuracy of the  inverse feature learning is just about $1\%$ below than $DrELM^{r}$ \cite{yu2015learning} and the accuracy for Segment data set is $2.2\%$ better than the known deep representation leanings as the-state-of-art in recent literature. %The proposed inverse feature learning as a preface to error representation learning that can be used alongside deep representation learning ones.

We should note that we evaluated the performance of this inverse feature learning method using other clustering techniques such as spectral clustering (e.g., DBSCAN \cite{ester1996density}). Using this clustering method led to better results since it can handle different data densities in better forms. However, this clustering works based on graph Laplacian matrix that requires a considerable memory to operate, therefore, it was not easily feasible for large scale data sets.

%\section{new}

%\begin{table}[t]
%\vspace{-10pt}
%\caption{The specification of data sets.}
%\label{datasetsdesriptoin}
%\vskip 0.15in
%\begin{center}
%\begin{small}
%\begin{sc}
%\begin{tabular}{lcccr}
%\toprule
%Data sets & \#instances & \#Dimension & \#class \\
%\midrule
%MNIST    & 70000 &  784& 10  \\
%Fasion-MNIST    & 70000 &  784& 10  \\
%USPS    & 9296 &  256& 10  \\
%REUTERS-10k    & 10000 &  2000& 4  \\
%\bottomrule
%\end{tabular}
%\end{sc}
%\end{small}
%\end{center}
%\vskip -0.1in
%\vspace{-12pt}
%\end{table}

\section{Conclusion} \label{sec:con}
In this paper, we propose error representation learning as a new feature learning trend that deals with error as a dynamic component that can disclose a valuable set of information about the relations of the instances and the classes. To the best of our knowledge, current machine learning methods interpret the error in a simple notion of a constant scalar that evaluates the differences between the true and the predicted labels. In this paper, we propose a general concept of error representation that can evaluate the error in several new levels in order to learn high-level and explicable features by trial. The proposed feature learning method based on error representation, called as \textit{inverse feature learning} adds the set of learned features to the set of primary features. 
The inverse feature learning method is performed in a hierarchical structure by evaluating the results of adding each instance of interest to available classes using several different metrics. The experimental results show the significant performance of this feature learning method compared to several state-of-the-art classification methods and some of the most known deep representation learning methods.

\bibliographystyle{IEEEtran}
%\bibliography{bibio}
\bibliography{refrences}

%\bibliographystyle{cas-model2-names}

%\bibliography{cas-refs}

\end{document}